\documentclass{llncs}
\usepackage{amssymb}
\usepackage{bm}
\usepackage{cite}
\usepackage{graphicx}
\usepackage{subfigure}
\usepackage{multirow}
\usepackage{amsmath}
\usepackage{graphicx}
\usepackage{tabularx}
\usepackage{url}
\urldef{\mailsa}\path|{alfred.hofmann, ursula.barth, ingrid.haas, frank.holzwarth,|
\urldef{\mailsb}\path|anna.kramer, leonie.kunz, christine.reiss, nicole.sator,|
\urldef{\mailsc}\path|erika.siebert-cole, peter.strasser, lncs}@springer.com|
\newcommand{\keywords}[1]{\par\addvspace\baselineskip
\noindent\keywordname\enspace\ignorespaces#1}
\usepackage{float}

\begin{document}

\mainmatter  

\title{Text Generation Based on Generative Adversarial Nets with Latent Variable}

\titlerunning{Text Generation Based on Generative Adversarial Nets with Latent Variable}
%
%
\author{ Heng Wang$^1$, Zengchang Qin$^1$$^*$, Tao Wan$^2$$^*$}
\authorrunning{Wang, Qin and Wan }
%
\institute{$^1$Intelligent Computing and Machine Learning Lab\\
     School of ASEE, Beihang University, Beijing, 100191, China\\
      $^2$School of Biological Science and Medical Engineering\\
      Beihang University, Beijing, 100191, China\\
     \tt  $^*$\{zcqin, taowan\}@buaa.edu.cn \\
}
%
%

\toctitle{Text Generation Based on Generative Adversarial Nets with Latent Variable}

\maketitle

\begin{abstract}
In this paper, we propose a model using generative adversarial net (GAN) to generate realistic text. Instead of using standard GAN, we combine variational autoencoder (VAE) with generative adversarial net. The use of high-level latent random variables is helpful to learn the data distribution and solve the problem that generative adversarial net always emits the similar data. We propose the VGAN model where the generative model is composed of recurrent neural network and VAE. The discriminative model is a convolutional neural network. We train the model via policy gradient. We apply the proposed model to the task of text generation and compare it to other recent neural network based models, such as recurrent neural network language model and SeqGAN. We evaluate the performance of the model by calculating negative log-likelihood and the BLEU score. We conduct experiments on three benchmark datasets,  and results show that our model outperforms other previous models.

\end{abstract}

\section{Introduction}
Automatic text generation is important in natural language processing and artificial intelligence. For example, text generation can help us write comments, weather reports and even poems. It is also essential  to machine translation, text summarization, question answering and dialogue system \cite{mikolov2010recurrent}. 
One popular approach for text generation is by modeling sequence via recurrent neural network (RNN) \cite{mikolov2010recurrent}.  However, recurrent neural network language model (RNNLM) suffers from two major drawbacks when used to generate text. First, RNN based model is always trained through maximum likelihood approach, which suffers from exposure bias \cite{bengio2015scheduled}. Second, the loss function used to train the model is at word level but the performance is typically evaluated at sentence level.
There are some research on using generative adversarial net (GAN) to solve the problems.
For example,
Yu \emph{et al.} \cite{yu2016seqgan:} applies GAN to discrete sequence generation by directly optimizing the discrete discriminator's rewards.
Li \emph{et al.} \cite{Li2017Adversarial} applies GAN to open-domain dialogue generation and generates higher-quality responses. Instead of directly optimizing the GAN objective, Che \emph{et al.} \cite{Che2017Maximum} derives a novel and low-variance objective using the discriminator's output that follows corresponds to the log-likelihood. 
In GAN, a discriminative net $D$ is learned to distinguish whether a given data instance is real or not, and a generative net $G$ is learned to confuse $D$ by generating highly realistic data. GAN has achieved a great success in computer vision tasks \cite{denton2015deep}, such as image style transfer \cite{gatys2015a}, super resolution and imagine generation \cite{radford2016unsupervised}.  
Unlike image data, text generation is inherently discrete, which makes the gradient from the discriminator difficult to back-propagate to the generator \cite{huszar2015how}.
Reinforcement learning is always used to optimize the model when GAN is applied to the task of text generation \cite{yu2016seqgan:}.


Although GAN can generate realistic texts, even poems, there is an obvious disadvantage that
GAN always emits similar data \cite{salimans2016improved}.
For text generation, GAN usually uses recurrent neural network as the generator. Recurrent neural network mainly contains two parts: the state transition and a mapping from the state to the output, and two parts are entirely deterministic. This could be insufficient to learn the distribution of highly-structured data, such as text \cite{chung2015a}. In order to learn generative models of sequences, we propose to use high-level latent random variables to model the observed variablity. We combine recurrent neural network with variational autoencoder (VAE) \cite{kingma2014auto-encoding} as generator $G$.

In this paper, we propose a generative model, called VGAN, by combining VAE and generative adversarial net to better learn the data distribution and generate various realistic text. The paper is structured as the following: In Section 2, we give the preliminary of this research. In Section 3, we introduce the VGAN model and adversarial training of VGAN. Experimental results are given and analyzed in Section 4. In Section 5, we conclude this research and discuss the possible future work.

\section{Preliminary}
\subsection{LSTM Architecture}
A recurrent neural network is a class of artificial neural network where connections between units form a directed cycle \cite{mikolov2010recurrent}. This allows it to exhibit dynamic temporal behavior. Long short-term memory (LSTM) is an improved version of recurrent neural network considering long-term dependency in order to overcome the vanishing gradient problem. It has been successfully applied  in many tasks, including text generation and speech recognition \cite{Hochreiter1997Long}. LSTM has a architecture consisting of a set of recurrently connected subnets, known as memory blocks. Each block contains memory cells and three gate units, including the input, output and forget gates. The gate units allow the network to learn when to forget previous information and when to update the memory cells given new information.

Given a vocabulary $V$ and an embedding matrix $W\in R^{{q}\times { |V|} }$  whose columns correspond to vectors; $|V|$ and $q$ denote the size of vocabulary  and the dimension of the token vector, respectively. The embedding matrix $W$ can be initialized randomly or pretrained. Let  $x$ denote a token with index $k$, $e(x) \in R^{|V| \times {1}}$ is a vector with zero in all positions except $R^{{|V| \times {1}}}_{k}=1$.
Given an input sequence $X_{s} = (x_{1},x_{2},\cdots,x_{T}) $, we compute the output sequence of LSTM $Y_{s} = (y_{1},y_{2},\cdots,y_{T}) $. When the input sequence passes through the embedding layer, each token is represented by a vector: $v_{i}= W \bm\cdot e(x_{i}) \in R^{q \times 1}$.  The relation between inputs, memory cells and outputs are defined by the following equations:
\begin{equation}
i^{(t)} = \sigma (W_{ix} v^{(t)} + W_{ih}  h^{(t-1)} + W_{ic}  c^{(t-1)})
\label{eq:pathnode}
\end{equation}
\begin{equation}
f^{(t)} = \sigma (W_{fx}  v^{(t)} +  W_{fh}  h^{(t-1)} + W_{fc}  c^{(t-1)})
\label{eq:pathnode}
\end{equation}
\begin{equation}
c^{(t)} = f^{(t)} \odot c^{(t-1)} + i^{(t)} \odot tanh(W_{cx}  v^{(t)} + W_{ch}  h^{(t-1)})
\label{eq:pathnode}
\end{equation}
\begin{equation}
o^{(t)} = \sigma (W_{ox}  v^{(t)} + W_{oh}  h^{(t-1)} + W_{oc}  c^{(t-1)})
\label{eq:pathnode}
\end{equation}
\begin{equation}
h^{(t)}=o_{(t)}\odot tanh(c^{t})
\label{eq:pathnode}
\end{equation}
where $i^{(t)}\in R^{l\times 1}$, $f^{(t)}\in R^{l\times 1}$, $o^{(t)}\in R^{l\times 1}$  and $h^{(t)}\in R^{l\times 1}$ represent the input gate, forget gate, output gate, memory cell activation vector and the recurrent hidden state at time step $t$; $l$ is the dimension of LSTM hidden units, $\sigma$ and $tanh$ are the logistic sigmoid function and hyperbolic tangent function, respectively. 
 $\odot$ represents element-wise multiplication \cite{Hochreiter1997Long}.

\subsection{Variational Autoencoder}
An autoencoder (AE) is an unsupervised learning neural network with the target values to be equal to the inputs. Typically, AE is mainly used to learn a representation for the input data, and extracts features and reduces dimensionality \cite{kingma2014auto-encoding}.
Recently, autoencoder has been widely used to be a generative model of image and text. The variational autoencoder is an improved version based on the standard autoencoder. For variational autoencoder, there is a hypothesis that data is generated by a directed model and some latent variables are introduced to capture the variations in the observed variables. The directed model $p(x)=\int_{}^{} p(x|z)p(z)\,dz$ is optimized by using a variational upper bound:
\begin{equation}
\begin{aligned}
-\log p(x) =&- \log \int_{}^{} p(x|z)p(z)\,dz \\
 \leq -KL(q(z|x)&||p(z)) + E_{z\sim q(z|x)}[\log p(x|z)]
\label{eq:pathnode}
 \end{aligned}
\end{equation}
where $p(z)$  is a prior distribution over the latent random variable $z$; The prior distribution is unknown, and we generally assume it to be a normal distribution.
$p(x|z)$ denotes a map from the latent variables $z$ to the observed variables $x$, and given $z$ it produces a distribution over the possible corresponding values of $x$.  $q(z|x)$ is a variational approximation of the true posterior distribution. let $KL(q(z|x)||p(z))$ denote the  Kullback-Leibler divergence between $q(z|x)$ and $p(z)$; The introduction of latent variables makes it intractable to optimize the model directly. We minimize the upper bound of the negative log-likelihood to optimize VAE. The training algorithm we use is Stochastic Gradient Variational Bayes (SGVB) proposed in  \cite{kingma2014auto-encoding}.

\subsection{Generative Adversarial Nets}

For generative adversarial nets, there is a two-sided  zero-sum  game between a generator and a discriminator.  The training objective of the discriminative model is to determine whether the data is from the fake data generated by the generative model or the real training data. For the generative model, its objective is to generate realistic data, which is similar to the true training data and the discriminative model can't distinguish. For the standard generative adversarial networks, we train the discriminative model $D$ to maximize the probability of giving the correct labels to both the samples from the generative model and training examples. We simultaneously train the generative model $G$ to minimize the estimated probability of being true by the discriminator. The objective function is:
\begin{equation}
\min_G \max_D V(D,G) = E_{x\sim p_{data}(x)}[\log D(x)] + E_{z\sim p_z(z)}[\log(1-D(G(z)))]
\label{eq:pathnode}
\end{equation}
where $z$ denotes the noises and $G(z)$ denotes the data generated by the generator.  $D(x)$ denotes the probability that  $x$ is from the training data with emperical distribution $p_{data}(x)$.

\section{Model Description}

\subsection{The Generative Model of VGAN}
\begin{figure}[!t]
\centering
\centerline{\includegraphics[scale=0.4]{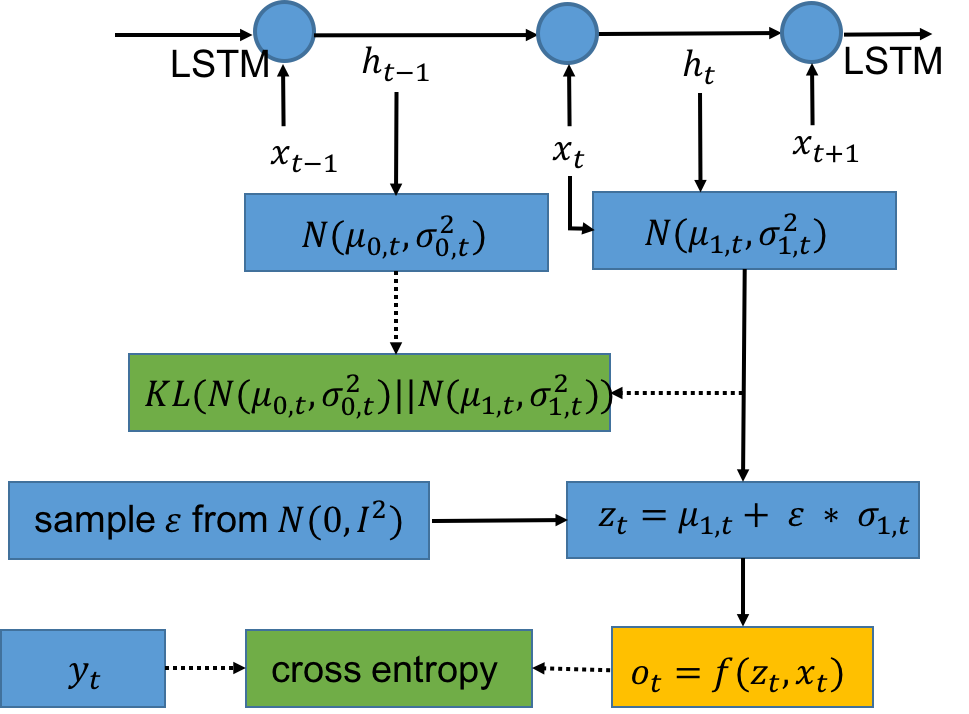}}
\caption{The structure of the generator $G_{\theta}$. The generator is composed of LSTM and VAE. $x_{t}$ denotes the input at timestep $t$; $h_{t}$ denotes the LSTM hidden state; $N(\mu_{0,t},\sigma_{0,t}^{2})$ denotes the prior distribution;  $N(\mu_{1,t},\sigma_{1,t}^{2})$ denotes the appriximate posterior distribution; $y_{t}$ denotes the target output at timestp $t$, which is the same as $x_{t+1}$.  $o_{t}$ denote the estimated result. The dotted line denotes the optimization process in the pre-training stage.}
\label{fig:fig1}
\end{figure}

	The proposed generative model contains a VAE at every timestep. For the standard VAE, its prior distribution is usually a given standard normal distribution. Unlike the standard VAE, the current prior distribution depends on the hidden state $h_{t-1}$ at the previous moment,  and adding the hidden state as an input is helpful to alleviate the long term dependency of sequential data. It also  takes consideration of the temporal structure of the sequential data \cite{chung2015a} \cite{serban2016a}. The model is described in Fig. \ref{fig:fig1}. The prior distribution $z_{t} = p_{1 }(z_{t}|h_{t-1})$ is:
%
\begin{equation}
z_{t} \sim  N(\mu _{0,t},\sigma ^{2} _{0,t}) ,   \quad   where \quad  [\mu_{0,t},\sigma_{0,t}^2] = \varphi^{prior}(h_{t-1})
\label{eq:pathnode}
\end{equation}
where $\mu_{0,t}$, $\sigma_{0,t}^2$ are the mean and variance of the prior Gaussian distribution, respectively. The posterior distribution depends on the current hidden state.
For the appriximate posterior distribution, it depends on the current state $h_{t}$ and the current input $x_{t}$: $z_{t}^{\prime}=q_{1}(z_{t}|x_{t},h_{t})$.
\begin{equation}
z_{t}^{\prime} \sim N(\mu _{1,t},\sigma ^{2} _{1,t}) ,   \quad   where \quad  [\mu_{1,t},\sigma_{1,t}^2] = \varphi^{posterior}(x_{t},h_{t})
\label{eq:pathnode}
\end{equation}
where $\mu_{1,t}$ , $\sigma_{1,t}^2$  are the mean and variance of the approximate posterior Gaussian distribution, respectively. $\varphi^{prior}$ and $\varphi^{posterior}$  can be any highly flexible functions, for example, a neural network.

	The derivation of the training criterion is done via stochastic gradient variational Bayes. We achieve the goal of minimizing the negative log-likelihood in the pre-training stage by minimizing $L(x_{1:T})$:
\begin{equation}
\begin{aligned}
L(x_{1:T}) = & -\log p(x_{1:T}) 
=  -\log \int_{z_{1:T}} \frac{q_{1}(z_{1:T}|x_{1:T},h_{1:T})}{q_{1}(z_{1:T}|x_{1:T},h_{1:T})} \prod_{t=0}^{T-1} p(x_{t+1}|x_{1:t},z_{1:t}) \,dz_{1:T} \\
 \leq & -KL(q_{1}(z_{1:T}| x_{1:T},h_{1:T}) || p_{1}(z_{1:T}|x_{1:T-1},h_{1:T-1})) 
 \\ + & E_{z_{1:T}  \sim q_{1}(z_{1:T}|x_{1:T},h_{1:T})} \left [  \sum_{t=0}^{T-1}\log p(x_{t+1}|x_{1:t},z_{1:t}) \right ] 
 \label{eq:pathnode}
\end{aligned}
\end{equation}
where $p_{1}$ and $q_{1}$ represent the prior distribution and the approximate posterior distribution.

	If we directly use the stochastic gradient descent algorithm to optimize the model, there will be a problem that some parameters of the VAE are not derivable. In order to solve the problem, we introduce the ``reparametrization trick" \cite{kingma2014auto-encoding}. For example, if we want to get the samples from the distribution $N(\mu _{1,t},\sigma ^{2} _{1,t})$, we will sample from a standard normal distribution $\epsilon \sim N(0,I^{2}) $  and get the samples $z_{t}^{\prime}$  via $z_{t}^{\prime} =\mu_{1,t} + \sigma_{1,t} \epsilon$.
Before the adversarial training, we need to pre-train the generative model via SGVB. For example, given the input $ X_{s} =( \textbf{S}, i, like, it )$, and the target output is $ Y_{s} =( i, like, it, \textbf{E} )$, where $\textbf{S}$ and $\textbf{E}$ are the start token and the end token of  a sentence, respectively. 
\begin{figure}[!t]
\centering
\centerline{\includegraphics[scale=0.4]{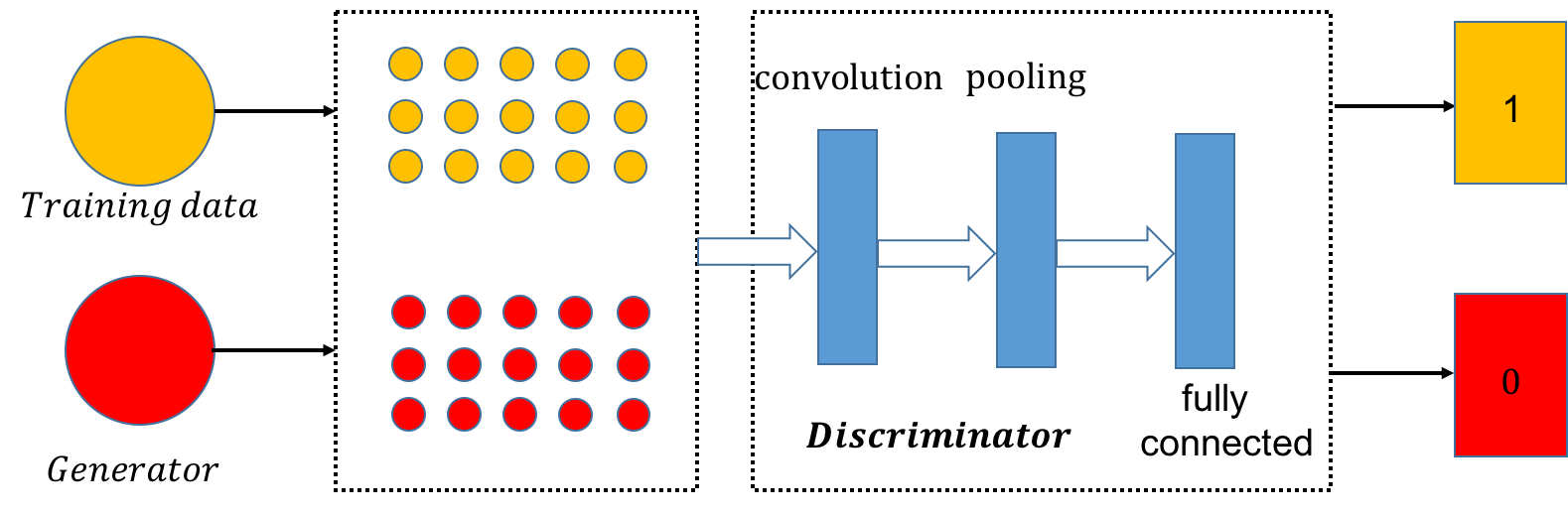}}
\caption{The illustration of the discriminator. The discriminator $D_{\phi}$ is trained by using the true training data and the fake data generated by generator. The discriminator contains the convolution layer, the max-pooling layer and the fully connected layer. }
\label{fig:fig2}
\end{figure}
\subsection{Adversarial Training of VGAN} 
\begin{figure}[!t]
\centering
\centerline{\includegraphics[scale=0.4]{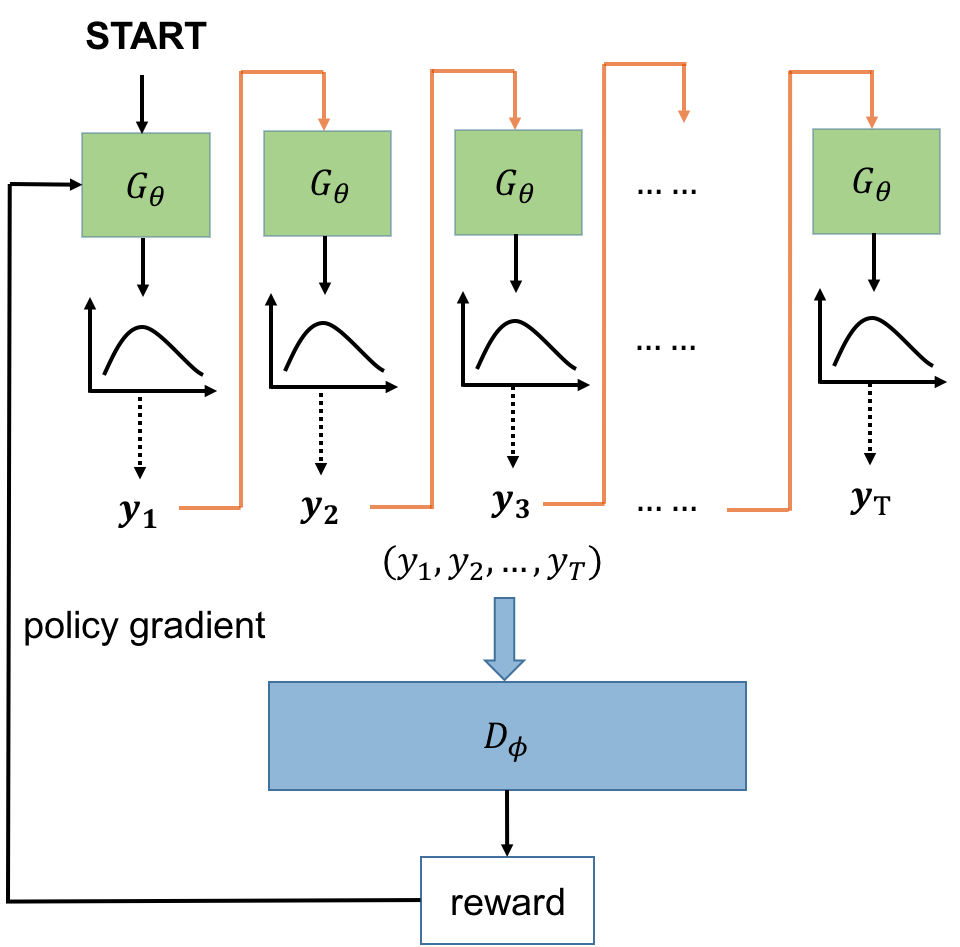}}
\caption{The training process of the generator via policy gradient. The dotted line denotes sampling a token from the output distribution. The sampled token is as the input at the next time. \textbf{START} denotes a start token.}
\label{fig:fig3}
\end{figure}

	In this paper, we choose convolutional neural network as the discriminative model, which has shown a great success in the task of text classification \cite{kim2014convolutional}\cite{kalchbrenner2014a}.
Let $v_{i} \in R^{k} $ denote a $k$-dimesion vector corresponding to the $i$-th word in the sentence.  $v_{1:n}=v_{1} \oplus v_{2} \oplus \cdots \oplus v_{n}$   denote a sentence of length $n$, where $\oplus$ is the concatenation operator and $v_{1:n} \in R^{n \times k} $  is a matrix. Then a filter $w_{1}\in R^{h \times k}$ is applied to a window of $h$ words to produce a new feature. For example, $c_{i}=f(w_{1} \bm\cdot v_{i:i+h-1} + b )$
 where $c_{i}$ is a feature generated by convolution operation. $f$ denotes a nonlinear function such as the hyperbolic tangent or sigmoid; $ b$ is a bias term. When the filter is applied to a sentence $v_{1:n}$,  a feature map $\mathbf{c}=[c_{1},c_{2},\cdots,c_{n-h+1}]$  is generated. We can use a variety of convolution kernels to obtain a variety of feature maps. We apply a max-over-time pooling operation to the features to get the maximum value $c' = \max \{\mathbf{c}\}$. Finally, all features are used as input to a fully connected layer for classification.
 After the generator is pre-trained, we use the generator to generate the negative examples. The negative samples and the true training data are combined as the input of the discriminator. The training process of discriminator is showed in Fig. \ref{fig:fig2}.
In the adversarial training, the generator $G_{\theta}$ is optimized via  policy gradient, which is a reinforcement learning algorithm \cite{sutton2000policy} \cite{ranzato2016sequence}.  The training process is showed in Fig. \ref{fig:fig3}. $G_{\theta}$ can be viewed as an $agent$, which interacts with the environment. The parameters $\theta$ of this $agent$ defines a policy, which determines the process of generating the sequences. 

Given a start token $ \textbf{S}$ as the input, $G_{\theta}$ samples a token from the generating distribution. And the sampled word is as the input at the next time. A whole sentence is generated word by word until an end token $\textbf{E}$ has been generated or the maximum length is reached.
For example, given a start token $\textbf{S}$ as the input, the sequence $Y_{1:T} = (y_{1},y_{2},\cdots,y_{T})$ is generated by $G_{\theta}$. In timestep $t$, the state $s_{t}$ is the current produced tokens $(y_{1},y_{2},\cdots,y_{t-1})$ and the action $a_{t}$ is to select the next token in the vocabulary.  After taking an action $a_{t}$, the agent updates the state $ (y_{1},y_{2},\cdots,y_{t})$. If the agent reaches the end,  the whole sequence has been generated and the reward will be assigned. During the training, we choose the next token according to the current policy and the current state. But there is a problem that we can observe the reward after a whole sequence. When $G_{\theta}$ generates the sequence, we actually care about the expected accumulative reward from start to end and not only the end reward. At every timestep, we consider not only the reward brought by the generated sequence, but also the future reward.

 In order to evaluate the reward for the intermediate state, Monte Carlo search has been employed to sample the remaining unknown tokens. In result, for the finished sequences, we can directly get the rewards by inputting them to the discriminator $D_{\phi}$. For the unfinished sequences, we first use the Monte Carlo search to get estimated rewards \cite{chaslot2008monte-carlo}. To reduce the variance and get more accurate assessment of the action value, we employ the Monte Carlo search for many times and get the average rewards. The objective of the generator model $G_{\theta}$ is to generate a sequence from the start token to maximize its expected end reward:
\begin{equation}
\max_{\theta} J(\theta) = E[R_{T}| s_{0}] = \sum_{y_1 \in V} G_{\theta}(y_1|s_0) \bm\cdot Q_{G_\theta}^{D_\phi}(s_0,y_1)
\label{eq:pathnode}
\end{equation}
where $R_{T}$ denotes the end reward after a whole sequence is generated; $Q_{G_\theta}^{D_\phi}(s_i,y_i)$ is the action-value function of a sequence, the expected accumulative reward starting from state $s_{i}$, taking action $a=y_{i}$; $G_{\theta}(y_{i}|s_{i})$ denotes the generator chooses the action $a=y_{i}$ when the state $s_{i}=(y_{1},y_{2},\cdots,y_{i-1})$ according to the policy. The gradient of the objective function $J(\theta)$ can be derived \cite{yu2016seqgan:} as:
\begin{equation}
\nabla_{\theta} J(\theta) = E_{Y_{1:t-1}\sim G_{\theta}} \left [\sum_{y_t \in V}  \nabla_{\theta }G_{\theta}(y_t|Y_{1:t-1}) \bm\cdot Q_{G_\theta}^{D_\phi}(Y_{1:t-1},y_t) \right ]
\label{eq:pathnode}
\end{equation}
we then update the parameters of $G_{\theta}$ as:
\begin{equation}
\theta \leftarrow \theta + \alpha \bm\cdot \nabla_{\theta} J(\theta) 
\label{eq:pathnode}
\end{equation}
where $\alpha$ is the learning rate.

After training the generator by using policy gradient, we will use the negative samples generated by the updated generator to re-train the discriminator $D_{\phi}$ as follows:
\begin{equation}
\min_{\phi} -E_{Y\sim p_{data}}[ \log D_{\phi}(Y)] - E_{Y\sim G_{\theta}}[ \log(1- D_{\phi}(Y))]
\label{eq:pathnode}
\end{equation}
The pseudo-code of the complete training process is shown in Algorithm \textbf{1}.
\begin{table}
\begin{tabular}{l}
\hline
\textbf{Algorithm 1.} Generative Adversarial Network with Latent Variable\\
\hline
\hline
\textbf{Require:} generaror: $G_{\theta}$; \\
\quad\quad discriminator: $D_{\phi}$; \\
\quad\quad True training dataset: $X$;\\
  1 :\quad Pre-train $G_{\theta}$ on $X$ by Eq. (10);\\
  2 :\quad Generate negative samples Y by using $G_{\theta}$;\\
  3 :\quad Pre-train $D_{\phi}$ on $X$ and $Y$ by minimizing the cross entropy;\\
  4 :\quad \textbf{Repeat:} \\
  5 :\quad\quad \textbf{for} 1 $\sim$ m \textbf{do}\\
  6 :\quad\quad\quad generate the data  $Y_{1}$ by using $G_{\theta}$;\\
  7 :\quad\quad\quad  use $D_{\phi}$ to get the reward of $Y_1$;\\
  8 :\quad\quad\quad  update the parameters $\theta$ of $G_{\theta}$ by Eq. (13); \\
  9 :\quad\quad \textbf{end for} \\
10:\quad\quad \textbf{for} 1 $\sim$ n \textbf{do}\\
11:\quad\quad\quad use $G_{\theta}$ to generate negative samples $Y_{2}$;\\
12:\quad\quad\quad update the parameters $\phi$ of $D_{\phi}$ on $X$ and  $Y_{2}$    by Eq. (14);\\
13:\quad\quad \textbf{end for} \\
14:\quad \textbf{Until:} \textbf{VGAN} converges. \\
\hline
\end{tabular}
\label{tab:AveSatis}
\end{table}

In this paper, we propose the VGAN model by combining VAE and GAN for modelling highly-structured sequences. The VAE can model complex multimodal distributions, which will help GAN to learn structured data distribution.

\section{Experimental Studies}
In our experiments, given a start token as input, we hope to generate many complete sentences. We train the proposed model on three datasets: Taobao Reviews, Amazon Food Reviews and PTB dataset \cite{marcus1993building}. Taobao dataset is crawled on taobao.com. The sentence numbers of Taobao  dataset, Amazon dataset are 400K and 300K, respectively.  We split the datasets into 90/10 for training and test. For PTB dataset is relatively small, the sentence numbers of training data and test data  are 42,068 and 3,370.
\subsection{Training Details}
In our paper, we compare our model to two other neural models RNNLM \cite{mikolov2010recurrent} and SeqGANs \cite{yu2016seqgan:}. For these two models, we use the random initialized word embeddings, and they are trained at the level of word. We use 300 for the dimension of LSTM hidden units. The size of latent variables is 60. For Taobao Reviews and PTB dataset, the sizes of vocabulary are both 5K. For Amazon Food Reviews, the size is 20K. The maximum lengths of Taobao Reviews, PTB dataset and Amazon Food Reviews are 20, 30 and 30, respectively. We drop the LSTM hidden state with the dropout rate 0.5. All models were trained with the Adam optimization algorithm \cite{kingma2015adam:} with the learning rate 0.001.
	First, we pre-train the generator and the discriminator. We pre-train the generator by minimizing the upper bound of the negative log-likelihood. We use the pre-trained generator to generate the negative data. The negative data and the true training data are combined to be the input of the discriminator. Then, we train the generator and discriminator iteratively. Given that the generator has more parameters and is more difficult to train than the discriminator, we perform three optimization steps for the discriminator for every five steps for the generator. The process is repeated until a given number of epochs is reached. 
\begin{table}[htbp]
\caption{BLEU-2 score on three benchmark datasets. The best results are highlighted.}
\label{table1}
\begin{center}
\begin{tabular}{c|c|c|c|c|c}
\hline
\hline
Numbers of generated sentence & \quad 200  \quad & \quad 400  \quad & \quad 600 \quad & \quad 800 \quad & \quad 1000\\
\hline
RNNLM (Taobao)  &\quad 0.965 \quad & \quad 0.967  \quad & \quad 0.967 \quad & \quad 0.967 \quad & \quad 0.967\\ 
\hline
SeqGAN (Taobao) &\quad 0.968 \quad & \quad 0.970 \quad & \quad 0.970 \quad & \quad 0.968 \quad & \quad 0.968 \\
\hline
VGAN-pre (Taobao) &\quad 0.968 \quad & \quad 0.968 \quad & \quad 0.967 \quad & \quad 0.968 \quad & \quad 0.968 \\
\hline
VGAN (Taobao) &\quad \textbf{0.969} \quad & \quad \textbf{0.972} \quad & \quad \textbf{0.970} \quad & \quad \textbf{0.969} \quad & \quad \textbf{0.969}\\
\hline
\hline
RNNLM (Amazon) &\quad 0.831 \quad & \quad 0.842  \quad & \quad 0.845 \quad & \quad 0.846 \quad & \quad 0.848\\ 
\hline
SeqGAN (Amazon)  &\quad 0.846 \quad & \quad 0.851 \quad & \quad 0.852 \quad & \quad 0.853 \quad & \quad 0.856 \\
\hline
VGAN-pre (Amazon) &\quad 0.842 \quad & \quad 0.849 \quad & \quad 0.854 \quad & \quad 0.849 \quad & \quad 0.848 \\
\hline
VGAN (Amazon) &\quad \textbf{0.876} \quad & \quad \textbf{0.874} \quad & \quad \textbf{0.866} \quad & \quad \textbf{0.868} \quad & \quad \textbf{0.868}\\
\hline
\hline

RNNLM (PTB) &\quad 0.658 \quad & \quad 0.650 \quad & \quad 0.654 \quad & \quad 0.655 \quad & \quad 0.662 \\
\hline
SeqGAN (PTB) &\quad 0.712 \quad & \quad 0.705  \quad & \quad 0.701 \quad & \quad 0.702 \quad & \quad 0.681\\ 
\hline
VGAN-pre (PTB) &\quad 0.680 \quad & \quad 0.690 \quad & \quad 0.694 \quad & \quad 0.695 \quad & \quad 0.671 \\
\hline
VGAN (PTB) &\quad \textbf{0.715} \quad & \quad \textbf{0.709} \quad & \quad \textbf{0.714} \quad & \quad \textbf{0.715} \quad & \quad \textbf{0.695}\\
\hline
\hline
\end{tabular}
\end{center}
\end{table}

\subsection{Results and Evaluation}
\begin{figure}
\centering
\subfigure[Taobao Reviews]{
\begin{minipage}[b]{0.31\linewidth}
\centering
\includegraphics[width=1.5in]{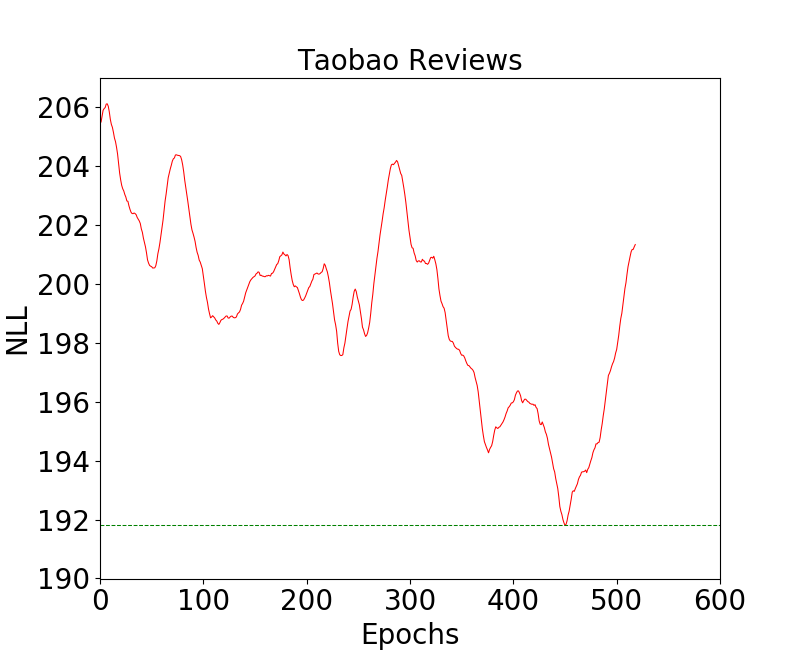}
\end{minipage}
}
\subfigure[Amazon Food Reviews]{
\begin{minipage}[b]{0.31\linewidth}
\centering
\includegraphics[width=1.5in]{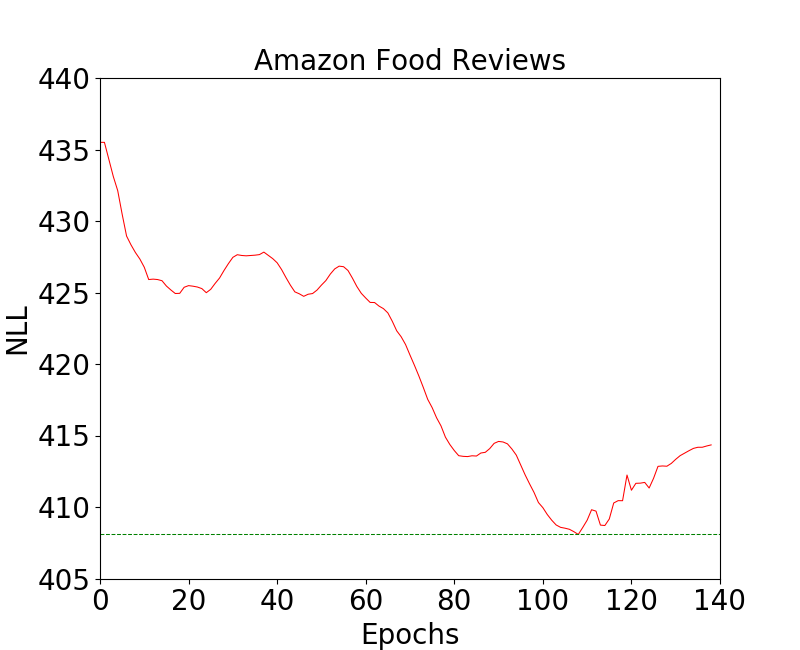}
\end{minipage}
}
\subfigure[PTB Data]{
\begin{minipage}[b]{0.31\linewidth}
\centering
\includegraphics[width=1.5in]{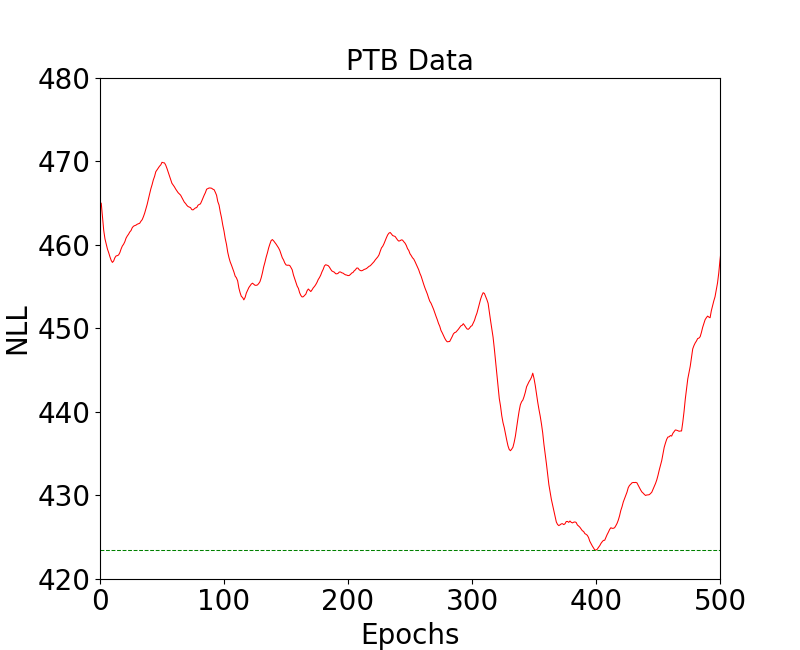}
\end{minipage}
}
\caption{(a) NLL values of Taobao Reviews. (b) NLL values of Amazon Food Reviews. (c) NLL values of PTB Data.}
\label{fig:fig 4}
\end{figure}

In this paper, we use the BLEU score \cite{papineni2002bleu:} and negative log-likelihood as the evaluation metrics. BLEU score is used to measure the similarity degree between the generated texts and the human-created texts. We use the whole test data as the references when calulating the BLEU score via nature language toolkit (NLTK). For negative log-likelihood, we calulate the value by inputting the test data. Table \ref{table2} shows the NLL values of the test data. VGAN-pre is the pretrained model of VGAN. 
\begin{equation}
NLL=-E \left [\sum_{t=1}^{T}\log G_{\theta}(y_t|Y_{1:t-1}) \right ]
\label{eq:pathnode}
\end{equation}
where $NLL$ denotes the negative log-likelihood;
Table \ref{table1} shows the experimental results of BLEU-2 score, and numbers in Table \ref{table1} denote the numbers of the sentence generated. We calulate the average BLEU-2 score between the generated sentences and the test data. The descent processes of NLL values during the adversarial training are showed in the Fig \ref{fig:fig 4}. Here, we give some examples generated by the proposed model. Due to the page limit, only some of generated comments from Amazon Food Reviews are shown in Table 3 and more results will be available online in the final version of the paper.
\begin{table}[htbp]
\caption{The comparison results (NLL) of VGAN to other models.}
\label{table2}
\begin{center}
\begin{tabular}{c|c|c|c}
\hline
Dataset  &\quad  Taobao Dataset \quad & \quad Amazon Dataset \quad & \quad PTB Dataset \\
\hline
RNNLM \cite{mikolov2010recurrent} &\quad 219 \quad & \quad 483 \quad & \quad 502 \\
\hline 
SeqGAN \cite{yu2016seqgan:} & \quad 212 \quad & \quad 467 \quad & \quad 490 \\
\hline 
VGAN-pre &\quad 205 \quad & \quad 435 \quad & \quad 465 \\
\hline 
VGAN  &\quad \textbf{191} \quad & \quad \textbf{408} \quad & \quad \textbf{423} \\
\hline
\end{tabular}
\end{center}
\end{table}

From Table \ref{table1} and Table \ref{table2}, we can see the significant advantage of VGAN over RNNLM and SeqGAN in both metrics. The results in the Fig \ref{fig:fig 4} indicate that applying adversarial training strategies to generator can breakthrough the limitation of generator and improve the effect.

\begin{table}
\label{table3}
\caption{Generated examples from Amazon Food Reviews using different models.}
\newcommand{\tabincell}[2]{\begin{tabular}{@{}#1@{}}#2\end{tabular}}
  \begin{tabular} {|l|l|}
  \hline
RNNLM & \tabincell{l}{ 1.  this is a great product , if you liked canned jerky but this is \\probably okay because your taste is great too .  \\ 2.  we 'll left never eating a bit of more bars . will definitely buy again .\\3.  but my friends and i love it mixes and they are hard in colored tap up ;\\ 4.  we love this product , and our purchase was fast too .\\ 5.  the soup is quite very good , its flax flavoring for the taste ... it is \\ pronounced \\ and yummy . \\} \\  
\hline
SeqGAN & \tabincell{l}{1.  how good it is that i 'll really cost again , this was my favorite .\\2.  service was super fast and good timely manner on each order .\\3.  the risk is very important , but truly the best so , you 'll \\ probably love it ! \\ 4.  each bag is n't lower and use .\\ 5.  i found that the seller was good to my own from amazon bags \\ practically  very frozen . \\} \\ 
\hline
\textbf{VGAN} & \tabincell{l}{1.  you just ate in first , but that is the best thing .\\  2.  but that did give me a much healthier and healthy .\\ 3.  the tea powder in need is based on the label and is very well as well .\\ 4.  chips ahoy cookies are very hard . this is just not very tasty . not what \\ they say ?\\ 5.  this is very nice and a little sweet . the red 's very fresh \\ 6.  i found that the coffee was not bold .\\ 7.  you are not sure why loves these cookies . i will be ordering these again \\ } \\  
\hline
\end{tabular}
\end{table}

\section{Conclusions}
In this paper, we proposed the VGAN model to generate realistic text based on classical GAN model. The generative model of VGAN combines generative adversarial nets with variational autoencoder and can be applied to the sequences of discrete tokens. In the process of training, we employ policy gradient to effectively train the generative model. Our results show that VGAN outperforms two strong baseline models for text generation and behaves well on three benchmark datasets. In the future, we plan to use deep deterministic policy gradient \cite{lillicrap2016continuous} to train the generator better. In the addition, we will choose other models as the discriminator such as recurrent convolutional neural network \cite{lai2015recurrent} and recurrent neural network \cite{liu2016recurrent}. 

\bibliographystyle{harvard}

\renewcommand\refname{References}

\
\end{document}